# Feature Fusion through Multitask CNN for Large-scale Remote Sensing Image Segmentation

Shihao Sun, Lei Yang, Wenjie Liu, Ruirui Li
Beijing University of Chemical Technology
College of Information Science & Technology
China
472527311@qq.com

*Abstract*—In recent years, Fully Convolutional Networks (FCN) has been widely used in various semantic segmentation tasks, including multi-modal remote sensing imagery. How to fuse multi-modal data to improve the segmentation performance has always been a research hotspot. In this paper, a novel end-to-end fully convolutional neural network is proposed for semantic segmentation of natural color, infrared imagery and Digital Surface Models (DSM). It is based on a modified DeepUNet and perform the segmentation in a multi-task way. The channels are clustered into groups and processed on different task pipelines. After a series of segmentation and fusion, their shared features and private features are successfully merged together. Experiment results show that the feature fusion network is efficient. And our approach achieves good performance in ISPRS Semantic Labeling Contest (2D).

## I. INTRODUCTION

With the development of remote sensing technology and the increasing demands for target recognition, a variety types of sensors have been invented to collect the image data. These sensors work on different wavelengths with different technical mechanism. It is a research hotspot that combining the heterogeneous remote sensing images to get dense, accurate semantic labeling for land cover or land use problems [1].

The deep neural network is initially designed for computer vision tasks on natural images of RGB channels [2][3]. Then the deep neural network has been successfully adopted to semantic segmentation on remote sensing imagery [4][5]. However, when facing multispectral or SAR images, these methods either roughly discard image channels except the RGB ones or coarsely fuse all the data to let the neural network make decision. Here, we propose an end-to-end fully convolutional neural network for semantic segmentation of heterogeneous remote sensing imagery. According to our approach, the data channels are firstly packed into different groups. Each group is assigned a processing pipeline for feature selection. Then all the features are concatenated together, followed by several convolutional layers. The weights of the features are computed by backward feedback propagating. In the ISPRS competition, compared with the method that roughly discards the channels, our approach gets a 2.6% higher accuracy improvement and compared with the method that coarsely fuses the data, our approach gets a higher accuracy improvement. Therefore, the multi-task fusion network could improve the efficiency of data usage. In summary, this paper has the following contributions:

1) The DeepUNet [6] is first adopted for semantic segmentation on very high resolution remote sensing images.

2) An information fusion method is proposed to deal with heterogeneous image segmentation in a multi-task way. It provides pixel-wise feature fusion and output pixel-wise segmentation map.

3) Compared with the DeepUNet trained by RGB image and that trained by mixed images, the multi-task feature fusion network got better performance.

## II. METHODOLOGY

### A. Overall architecture

We propose a multi-task approach that takes full advantage of the information from different sensors to process the images. In the preprocessing step, the diversity of the channel pairs is analyzed through (1) in which p, q are image channels, p(x) means the value of a pixel x in the channel p.

$$H(p, q) = \sum_{x} p(x) \cdot \log \left( \frac{1}{q(x)} \right) \quad (1)$$

TABLE I. THE CROSS ENTROPY OF DIFFERENT CHANNELS. THE PIXEL VALUE IS IN [0, 255].

| $\times 10^{10}$ | IR) | R) | G) | B) | DSM) |
|---|---|---|---|---|---|
| H(IR, | \ | -1.75 | -1.77 | -1.73 | -1.46 |
| H(R, | -1.54 | \ | -1.56 | -1.53 | -1.27 |
| H(G, | -1.68 | -1.65 | \ | -1.64 | -1.36 |
| H(B, | -1.55 | -1.51 | -1.54 | \ | -1.25 |
| H(DSM, | -0.75 | -0.73 | -0.74 | -0.72 | \ |

On the ISPRS postdam 2D dataset, there are R, G, B, IR and DSM five channels. The channels are packed into groups according to the diversity in an unsupervised learning way. In computer vision, three channels pictures are often used. So we also combine the three channels with the smallest cross entropy. The calculation results are shown in TABLE I, in which two results with larger absolute values are marked in blue. We got three groups: IRRG, IRGB, DSM (it has different orders of magnitude with others, so it alone as a group). For the training step, we design an end-to-end neural network. It includes several segmentation pipelines. Each group of data is assigned a task and input to the segmentation pipeline. After segmentation of DeepUNet, we concatenate three results and put them into a 1×1 convolution layer. This is to extract the shared features between them. Then, the shallow features of each groups are concatenated with the result respectively, and put into a 1×1 convolution layer to get the private features. In the end, fuse these features by a 1×1 convolution layer and a softmax layer outputs the final result. Fig. 1 depicts the architecture of our method.

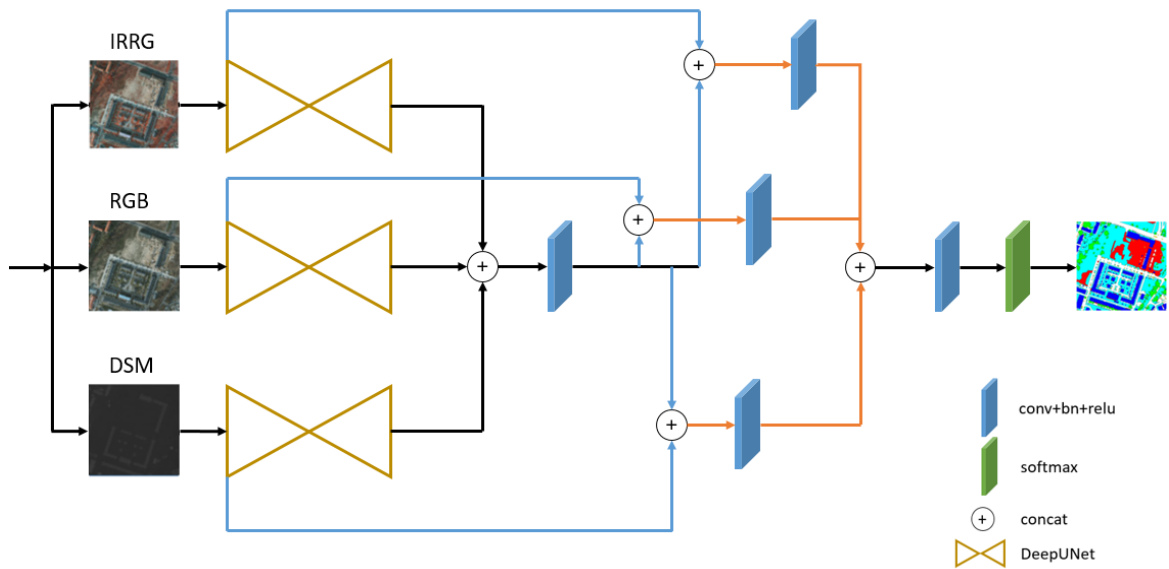

Fig. 1. The architecture of our method.

## B. DeepUNet segmentation pipeline

The DeepUNet is one of our previously proposed neural network for sea-land segmentation. Because of its success for VHR image segmentation, it is chosen to finish the semantic segmentation task on each pipeline. Its structure is shown in Fig. 2.

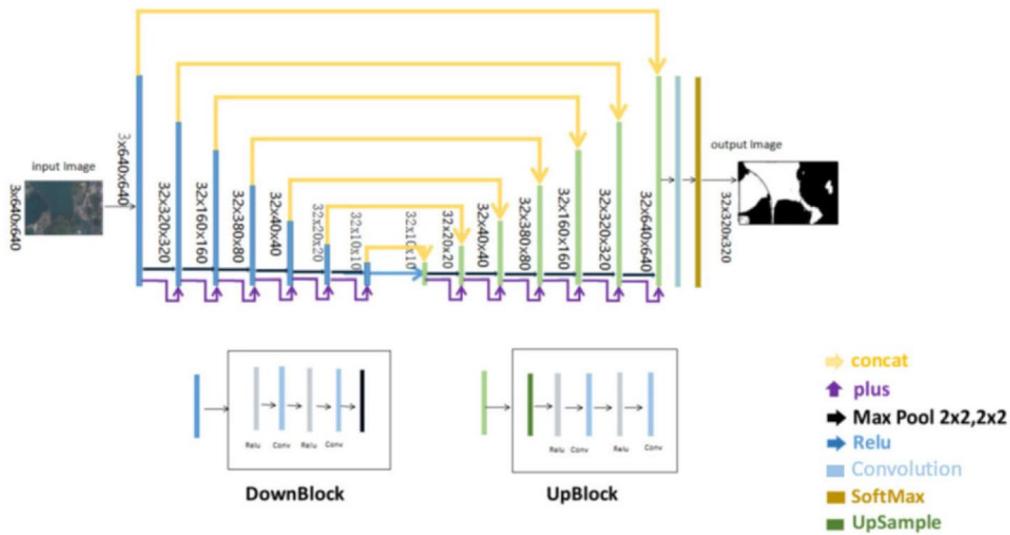

Fig. 2. The structure of DeepUNet.

The DownBlock is composed of two convolution layers and a max-pooling layer. The first convolution layer has 3×3 kernel size, 1×1 stride and 64 cores. The second convolution layer has 3×3 kernel size, 1×1 stride and 32 cores. The max-pooling layer has 2×2 kernel size and 2×2 stride. The input of max-pooling layer is the input of DownBlock plus the output of second convolution layer. This result is not only passed to the max-pooling layer, but also concatenated the feature maps to the corresponding UpBlock. The UpBlock also contains two convolution layers and an upsample layer. The input of first convolution layer concatenates two kinds of feature maps, the output of last UpBlock and the corresponding DownBlock's feature maps.

## III. EXPERIMENT

To show the effectiveness of our approach, we carry out comprehensive experiments on ISPRS dataset.

The experiments are carried out on a laboratory computer. Its configuration is shown in TABLE II. The operating system is installed of Ubuntu 16.04. The main required packages include python 2.7, CUDA8.0, cudnn7, mxNet0.10.0 and etc.

TABLE II. EXPERIMENTAL ENVIRONMENT

| CPU | Intel (R) Core (TM) i7-4790K 4.00Hz |
|---|---|
| GPU | GeForce GTX1080 Ti |
| RAM | 20GB |
| Hard disk | Toshiba SSD 512G |
| System | Ubuntu 16.04 |

The dataset is provided by ISPRS Semantic Labeling Contest (2D). The dataset contains 38 patches (of the same size), each consisting of a true orthophoto (TOP) extracted from a larger TOP mosaic. The dataset has been classified manually into six most common land cover classes. It provides 24 classification data (label images), while the ground truth of the remaining scenes will remain unreleased and stays with the benchmark test organizers to be used for evaluation of submitted results.

There is an image has some errors in the 24 label images. So, we only use 18 images for training, 5 images for validating. Data augmentation is essential to teach the network the desired invariance and robustness properties when only few training samples are available. We use rotation and scale variations as major method to augment the training set.

TABLE III. THE OVERALL ACCURACY OF COMPARED METHODS

| only RGB images | all channels mixed | proposed method |
|---|---|---|
| 88.0% | 88.4% | 90.6% |

In the same experimental environment, we compare the feature fusion method with the other two methods, that is, only using the RGB image or using the all channels image. TABLE III shows the accuracy of whether use fusion method. It can be found that the accuracy of our method is improved by 2.6% over only using the RGB image, and 2.2% more than with the all channels image.

TABLE IV is the result of our fusion method compared with other fusion method.

TABLE IV. PART OF RESULTS ON ISPRS 2D POTSDAM DATASET

|  | Imp surf | Building | Low veg | Tree | Car | Overall acc |
|---|---|---|---|---|---|---|
| DST_5 | 92.5% | 96.4% | 86.7% | 88.0% | 94.7% | 90.3% |
| ours | 93.2% | 96.7% | 87.1% | 88.6% | 95.9% | 90.6% |

A visualization result is shown in Fig. 3. The first column is original and resized true orthoimage for better comparison. Fig. 3 (b) is the segmented and classified result of our information fusion method. The last column is red/green image, indicating wrongly classified pixels.

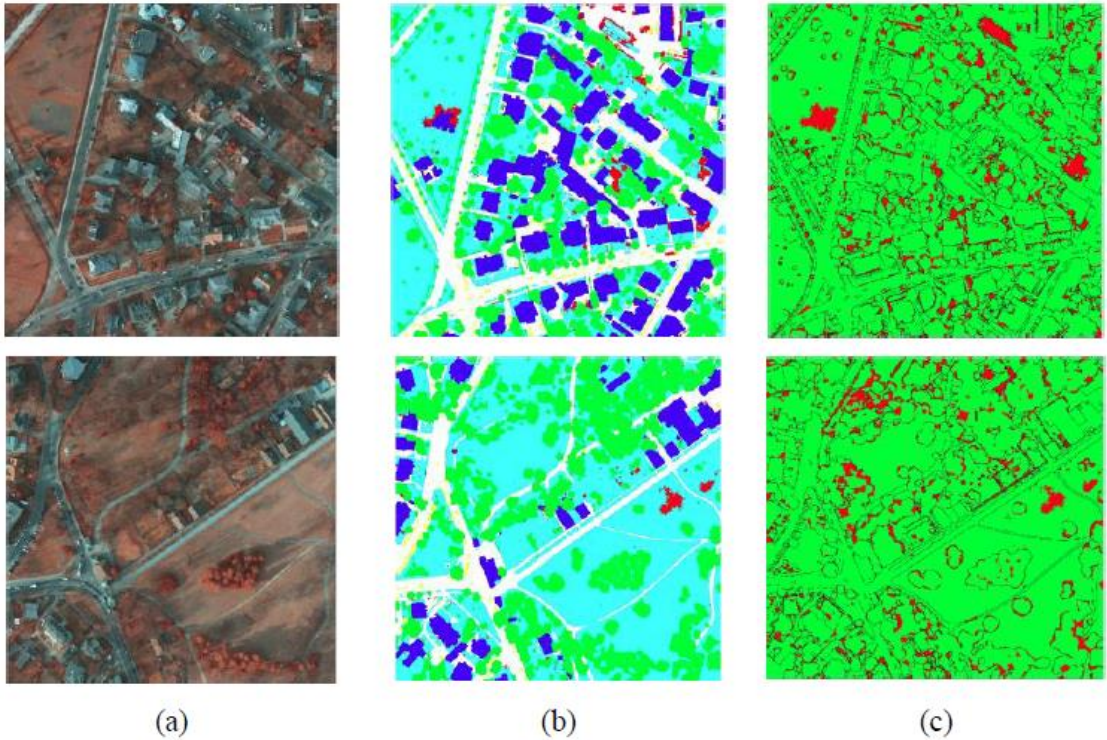

(a)       (b)       (c)

## IV. CONCLUSION

In the ISPRS Semantic Labeling Contest (2D), the accuracy of the method we proposed can be ranked in the first three (more detailed results can be found at http://www2.isprs.org/commissions/comm2/wg4/potsdam-2d-semantic-labeling.html, BUCT_3). The results show the effectiveness of our method.